\pdfoutput=1

\documentclass[letterpaper, 10 pt, journal, twoside]{IEEEtran} 

\usepackage{amsmath} 
\usepackage{amssymb}  
\usepackage[pdftex]{graphicx}
\usepackage{cite}
\usepackage{fancyhdr}

\begin{document}
\title{Learning Free Gait Transition for Quadruped Robots via Phase-Guided Controller}

\author{Yecheng Shao$^{1,2,3,4}$, Yongbin Jin$^{1,2,3,4}$, Xianwei Liu$^{1,2,3}$, Weiyan He$^{1,2}$, Hongtao Wang$^{1,2,3,4}$, Wei Yang$^{1,3}$%
\thanks{Manuscript received: September, 7, 2021; Revised November, 19, 2021; Accepted December, 8, 2021.}
\thanks{This paper was recommended for publication by Editor Jens Kober upon evaluation of the Associate Editor and Reviewers' comments. 
	This work was supported by the State Key Laboratory of Fluid Power and Mechatronic Systems (Zhejiang University).}
\thanks{$^{1}$All authors are with the Center for X-Mechanics, Zhejiang University, Hangzhou 310027, China
		{\tt\small \{shaoyecheng, yongbinjin, 3140104175, heweiyan, htw, yangw\}@zju.edu.cn}}%
\thanks{$^{2}$Y. Shao, Y. Jin and H. Wang are with the ZJU-Hangzhou Global Scientific and Technological Innovation Center, Hangzhou 310027, China}%
\thanks{$^{3}$Y. Shao, Y. Jin, X. liu, W. He and H. Wang are with the State Key Laboratory of Fluid Power and Mechatronic System, Zhejiang University, Hangzhou 310027, China}%
\thanks{$^{4}$Y. Shao, Y. Jin, X. Liu, H. Wang and W. Yang are with the Institute of Applied Mechanics, Zhejiang University, Hangzhou 310027, China}%
\thanks{Digital Object Identifier (DOI): 10.1109/LRA.2021.3136645}
}

\markboth{IEEE Robotics and Automation Letters. Preprint Version. Accepted December, 2021}
{Shao \MakeLowercase{\textit{et al.}}: Learning Free Gait Transition for Quadruped Robots via Phase-Guided Controller} 

\maketitle
\thispagestyle{fancy}
\fancyhf{}
\renewcommand{\headrulewidth}{0pt}
\fancyhead[LO]{\fontsize{8}{10}\selectfont \copyright 2021 IEEE.  Personal use of this material is permitted.  Permission from IEEE must be obtained for all other uses, in any current or future media, including reprinting/republishing this material for advertising or promotional purposes, creating new collective works, for resale or redistribution to servers or lists, or reuse of any copyrighted component of this work in other works.}

\begin{abstract}
Gaits and transitions are key components in legged locomotion. For legged robots, describing and reproducing gaits as well as transitions remain longstanding challenges. Reinforcement learning has become a powerful tool to formulate controllers for legged robots. Learning multiple gaits and transitions, nevertheless, is related to the multi-task learning problems. In this work, we present a novel framework for training a simple control policy for a quadruped robot to locomote in various gaits. Four independent phases are used as the interface between the gait generator and the control policy, which characterizes the movement of four feet. Guided by the phases, the quadruped robot is able to locomote according to the generated gaits, such as walk, trot, pacing and bounding, and to make transitions among those gaits. More general phases can be used to generate complex gaits, such as mixed rhythmic dancing.  With the control policy, the Black Panther robot, a medium-dog-sized quadruped robot, can perform all learned motor skills while following the velocity commands smoothly and robustly in natural environment.
\end{abstract}


\section{INTRODUCTION}
\IEEEPARstart{A}{nimals} can select the most suitable gait and make transitions according to the speed and terrains to traverse various environments in a robust and agile manner\cite{Hoyt1981,Drew2004}. To reproduce the agile locomotion, one key step is to reproduce each gait and all the transitions with a robust controller for quadruped robots. The state-of-the-art controllers, including model-based controllers\cite{8594448,kim2019highly,10.3389/frobt.2020.528473,gaertner2021collisionfree} and reinforcement learning (RL) controllers\cite{Hwangboeaau5872,Leeeabc5986,gangapurwala2020rloc,tan2018simtoreal}, have achieved excellent performance in outdoor locomotion tests with specific gaits. Both types of controllers, however, can hardly make free transitions according to speed or terrains.

Model based methods, such as the model predictive control (MPC), requires online optimization with simplified dynamic model. The performance is limited by the trade-off of model accuracy and control frequency. To achieve real-time online optimization, the contact sequence is predefined, which consequently determines the gait and the transition.\cite{8594448,kim2019highly} On the other hand, RL controllers can learn the most suitable locomotion. Achieving desired gaits and transitions from scratch requires well-designed reward functions, which is in general laborious. Motion imitation with RL offers an efficient way to build controllers to carry out similar behaviors to reference motions while satisfying constraints and achieving goals.\cite{2018-TOG-deepMimic} However, it is impractical to sample all feasible transitions starting from different states. In addition, due to the task-specific feature of machine learning, learning multiple gaits and transitions is a typical problem of multi-task learning, which remains challenging. It is noted that gait transition is closely related to foot sequence. The above difficulties can be overcome by introducing a central pattern generator (CPG). Smooth evolution between two given states in phase portrait can be generated, representing the complex behaviors of gait transitions.\cite{IJSPEERT2008642,Fukuoka2015,10.3389/fnbot.2016.00006} In this way, the multi-task learning problem is transformed into directly mapping CPG signals to foot trajectories while remaining dynamic balance.

In this work, we propose a phase-guided reinforcement learning framework for a quadruped locomotion controller. A set of four phases are added to the input of the RL controller. The mapping from the phase sets to the four foot trajectories can be constructed by imitation reinforcement learning. By training the RL controller with various phase-gait pairs, the quadruped robot can reproduce any learnt complex gaits and transitions. Our first contribution is the phase-guided reinforcement learning framework that minimizes the efforts in multitask learning by transforming the task of learning multiple gaits to learning the relation between phases and foot movements. We use phases to parameterize various gaits as the reference motion for the motion imitation tasks. The framework is flexible and compatible with different gait generators, including common generators like CPG and manually designed functions. We show that the policy can learn those gaits in a parameterized way by the sensitivity analysis of the neural network. Our second contribution is the demonstration that, under this framework, the CPG can be effectively used in guiding various gaits, such as walk, trot, pacing, and bounding, and free transitions from one to the other. All learned gaits and transitions in simulation can be reproduced on the hardware successfully.

\section{RELATED WORKS}
\subsection{Central Pattern Generators in Legged Robotics}
In biology, central pattern generators are biological neural networks which can produce rhythmic signals to body without rhythmic input\cite{IJSPEERT2008642}. It is generally considered as the source of periodic and structured motions like locomotion. In robotics, CPG is commonly regarded as a mathematical framework to formulate periodic patterns. As shown in \cite{dutta2019,Righetti2006}, well-designed CPGs are promising tools to represent any gaits and transitions. By analogy, it is natural to relate the CPG outputs to the joint commands in robot locomotion. The balance can be achieved with manually crafted feedback. Those works mainly focus on the structure of CPG network\cite{habu2019}, the mathematical forms\cite{10.3389/fnbot.2016.00006,dutta2019,app8010056}, the design of feedback control\cite{6696353,4543306} and the ability to transit among various gaits\cite{Fukuoka2015,7090642}. Unfortunately, none of the above works have demonstrated agile and robust locomotion with quadruped robots.

\subsection{Reference-Based RL Controllers for Legged Robotics}
Reinforcement learning has been demonstrated to be a promising tool to build controllers for legged robotics\cite{Hwangboeaau5872,Leeeabc5986,gangapurwala2020rloc,tan2018simtoreal}. With well-designed reward functions, controller for locomotion can be trained from simulation and deployed to hardware\cite{Hwangboeaau5872}. Designing reward functions often require tedious labor work. Motion imitation is a promising method to combine prior knowledge with RL and to reduce the work of reward-shaping. DeepMimic\cite{2018-TOG-deepMimic} is an inspiring approach that trains the policy to imitate the reference motion while satisfying physical constraints. For legged robotics, references from motion-capture, human sketch, and model-based controllers can all serve as good references to train controllers\cite{RoboImitationPeng20,siekmann2020learning,Li2021}. Specially, Li et al.\cite{Li2021} parameterized different reference motion from a HZD controller and show that the learned controller can handle more situations than the HZD controller.

\subsection{Multi-task Learning in Legged Locomotion}
The above motion imitation tasks only learn one type of motion. To overcome this issue, methods like multiplicative compositional policies\cite{MCPPeng19} and latent space models\cite{CARL2020} can be used to learn multiple motions primitives and mix them. Mix-of-experts (MoE) architecture has also been employed to handle the complexity of quadruped gaits and even fall-recovery tasks\cite{Yangeabb2174}.

Integrating task-specific signals into the observation space is another solution for learning multiple gaits. A recent work from Siekmann \textit{et al.}\cite{siekmann2021simtoreal} have shown bipedal controller for multiple gaits by using a cycle time with offsets and a vector ratio as gait representation and periodic reward for training. Compared to naive training approaches, the design of the contact-related periodic reward can capture the major characters of gaits. However, it is still not sufficient to avoid undesired behaviors. The task-specific settings are required in training. It is also noted that the dimension of the vector ratio $\boldsymbol{r}$ varies with different gaits, \textit{e.g.} $r\in\mathbb{R}$ for 1-beat gaits like hopping, $\boldsymbol{r}\in\mathbb{R}^{2}$ for 2-beats gaits like walking and $\boldsymbol{r}\in\mathbb{R}^{4}$ for 4-beats gaits like skipping. This will lead to different NN structures for different gaits. Reske \textit{et al.}\cite{reske2021imitation} include the origin phase in the input and use the MPC-Net\cite{Carius_2020} strategy to train the MoE network policy with multiple gaits. As the training dataset is generated using an MPC controller, the RL controller mostly reproduces the MPC performance.

\section{PHASE-GUIDED CONTROLLER}
\subsection{Overview}
\begin{figure}[!t]
	\centering
	\includegraphics[width=3.4in]{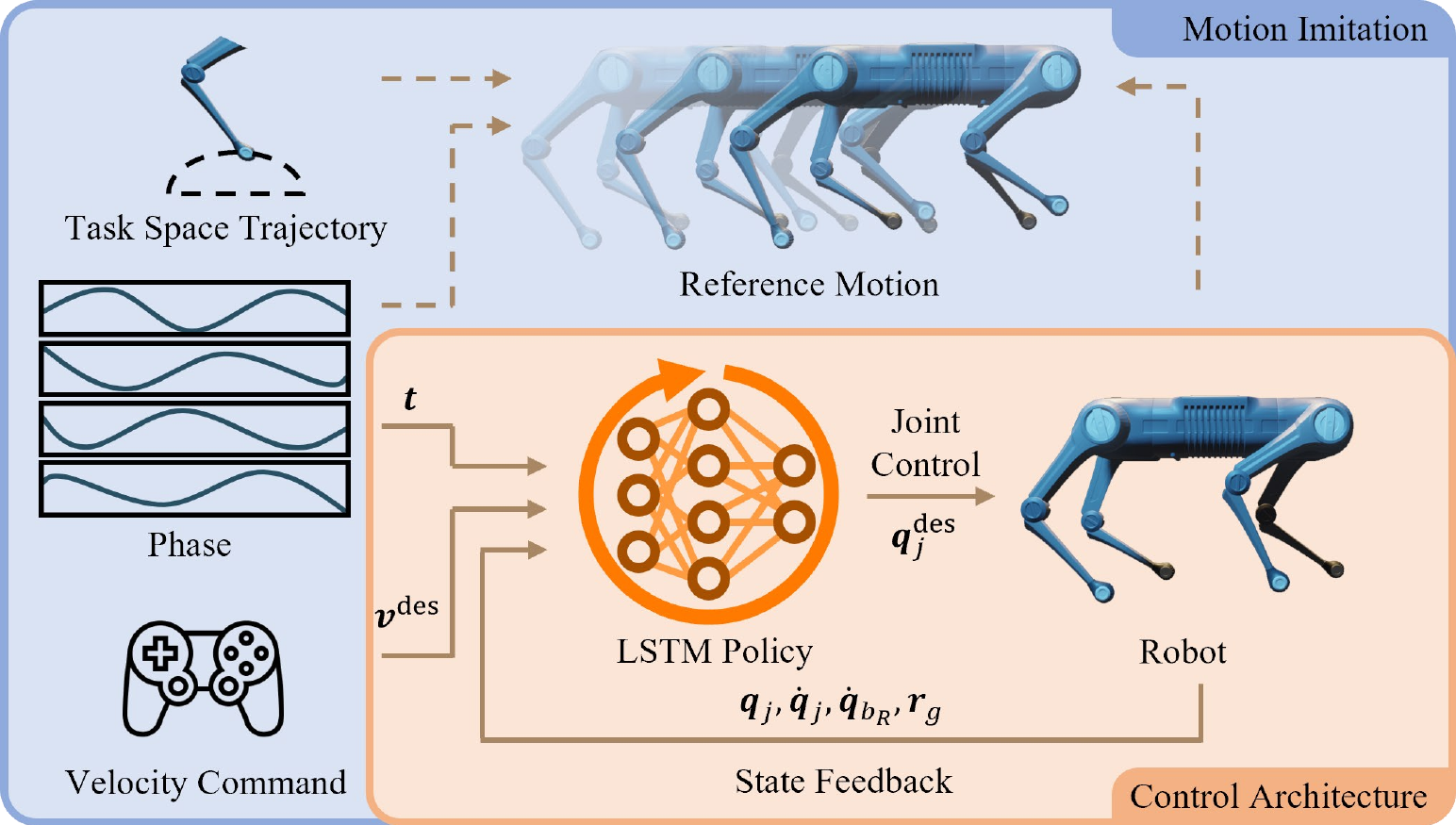}
	\caption{A flow chart of the phase-guided reinforcement learning framework. A set of four phases, representing four foot sequence, are used to generate the reference motions, which guides the control policy training. With motion imitation learning, the controller is able to learn gaits including walk, trot, pacing, bounding and transitions among those gaits.}
	\label{method_1}
\end{figure}

Fig. \ref{method_1} shows the phase-guided reinforcement learning framework. A set of four phases represent four foot sequences. The corresponding reference motions are generated by combining phases with velocity command and task space trajectories. The policy is trained to imitate reference motions while retaining balance and implementing velocity commands. For the purpose of locomotion control, the policy takes velocity commands, phases, and the state of robot as inputs and desired joint positions as outputs. 

\subsection{Motion Synthesis}
\begin{figure}[!t]
	\centering
	\includegraphics[width=3.4in]{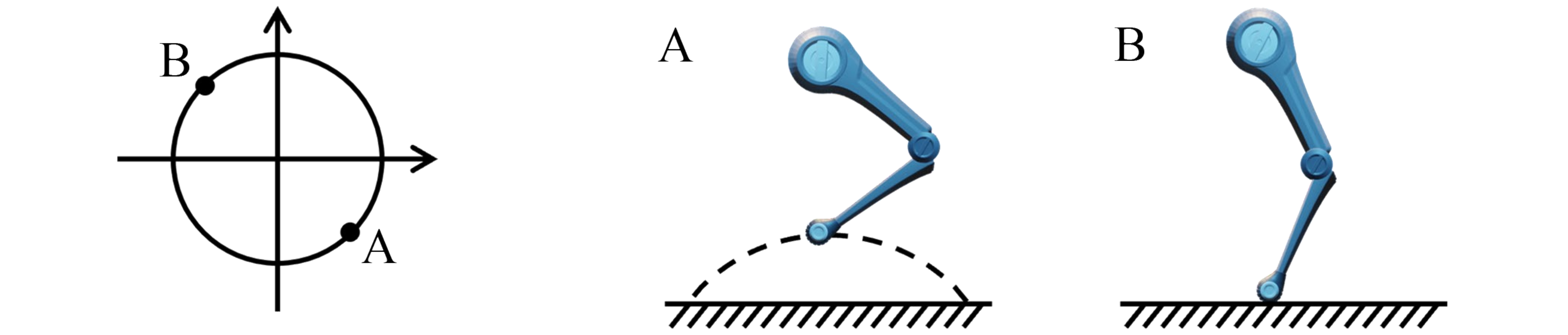}
	\caption{Illustration for the mapping from phase to leg movements. Point A is an example of the swing state and point B the stance state.}
	\label{method_2}
\end{figure}

\begin{figure}[!t]
	\centering
	\includegraphics[width=3.4in]{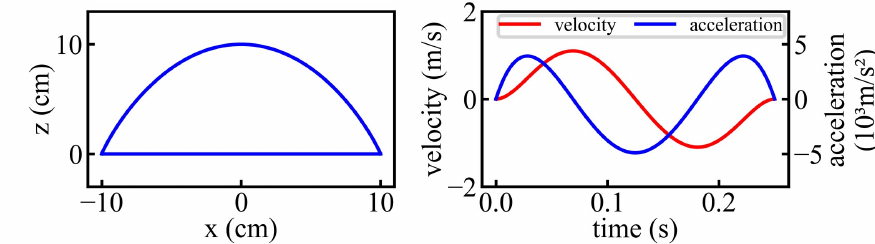}
	\caption{Illustration for the foot trajectory. The left figure shows the whole trajectory relative to the initial position in the horizontal frame, while the right figure shows the velocity and acceleration.}
	\label{method_3}
\end{figure}

Reference motion is one key part in motion imitation. To ensure the sample efficiency, manually synthesized motions, instead of captured motion, are used in this work. Generally, any quadruped locomotion can be simplified and decomposed into the periodic motions of four legs. Fig. \ref{method_2} illustrates the motion generation strategy based on four periodic leg phases. The four periodic phase variables, denoted as $\varphi_i\in[-\pi,\pi), \left(i=1,2,3,4\right)$, decrease over time, which is equivalent to the clockwise motion along the unit circle. The stance state of the i-th leg is represented by $\varphi_i\in[0,\pi)$ and the swing state by $\varphi_i\in[-\pi,0)$. The foot trajectory, as determined by both phases and desired velocities, is modelled by polynomials in (\ref{eq_method_traj}).
\begin{eqnarray}
	\label{eq_method_traj}
	\begin{aligned}
		p_i &=
		\begin{cases}
			-\frac{2\varphi_i}{\pi}-1,&\varphi_i\in[-\pi,0) \\
			\frac{2\varphi_i}{\pi}-1,&\varphi_i\in[0,\pi)
		\end{cases} \\
		r^x_i &= a_i^x\left(6p_i^5-15p_i^4+10p_i^3-0.5\right) \\
		r^y_i &= a_i^y\left(6p_i^5-15p_i^4+10p_i^3-0.5\right) \\
		r^z_i &= 
		\begin{cases}
			0,&stance \\
			h(-64p_i^6+192p_i^5-192p_i^4+64p_i^3),&swing
		\end{cases}
	\end{aligned}
\end{eqnarray}
Here, we introduce $p_i$ as an intermediate variable to represent phases at different states. The foot positions are denoted as $\left(r^x_i, r^y_i, r^z_i\right)$ with $i=1,2,3,4$. The coordinate framework for those positions is defined in the horizontal frame relative to the initial position of each foot. The $x$ and $y$ axes are the projections of forward and lateral axes of the body frame on the horizontal plane and the $z$ axis is upward vertical. Given the desired forward velocity $v_x$, lateral velocity $v_y$ and the yaw velocity $\omega$ in the horizontal plane, the step lengths in forward and lateral direction are calculated as $a_i^x=v_x T\beta, a_i^y=(v_y+k_i \omega l_x/2)T\beta$, respectively, in which $T$ is the period, $\beta$ the duty factor representing the fraction of stance phase in the whole period, $l_x$ the body length and $k_1=k_2=1, k_3=k_4=-1$. The yaw velocity is realized by the lateral velocity difference between the front and hind legs. The swing height is fixed at $h=8cm$. This polynomial is designed to ensure the first and second order derivatives at $\varphi=0$ and $\varphi=\pi$ to be zero, as shown in Fig. \ref{method_3}, corresponding to zero velocity and acceleration when the foot touches or leaves the ground. Joint positions and velocities are obtained through analytic inverse kinematics (IK). The design of using four independent phases guarantees the freedom in motion synthesis compared to using clock and offsets. Our unified representation is capable to generate rhythmic gaits without any modification.

\subsection{Gaits and Phases}\label{sec_gait}
Phase generation is an independent module and can be implemented in different ways. In this work, we use both CPGs and manually designed functions for demonstration, showing the generality of the framework. The CPG works well for common gaits as well as gait transitions, while manually designed functions are more flexible and can be used for complex rhythmic gaits.

We use the Hopf oscillator\cite{7090642} to construct the CPG. The modified Hopf oscillator formulates a planar dynamics system with a limit circle in the following form:
\begin{eqnarray}
	\label{eq_method_hopf_0}
	\begin{cases}
		\dot{\rho_0} &=\alpha\left(\mu^2-\rho^2\right)\rho_0+\gamma\rho_1 \\
		\dot{\rho_1} &=\alpha\left(\mu^2-\rho^2\right)\rho_1-\gamma\rho_0 \\
		\gamma &=\frac{\pi}{\beta T \left(e^{-b\rho_1}+1\right)} + \frac{\pi}{\left(1-\beta\right) T \left(e^{b\rho_1}+1\right)}
	\end{cases}
\end{eqnarray}
in which $\rho_0=\rho\cos\varphi$ and $\rho_1=\rho\sin\varphi$, $\varphi$ is the phase of oscillator, $\mu$ the radius of the limit circle and $\gamma$ the angular velocity for the oscillator on the limit circle. The parameters $\alpha, \beta, b$ and $T$ characterized the behaviors of the oscillator. The parameter $\alpha$ determines the speed for the oscillator to converge to the limit circle, which must be positive. The positive constant $b$ serves to ensure the value $\mu$ varies smoothly across different half planes. The period of limit circle is denoted as $T$, corresponding to the period of the motion of the leg. The duty factor $\beta$ determines the fraction of stance phase in a whole period. In practice, the parameter $\alpha$, and $b$ are fixed while $\beta,T$ are used for the gait modulation. Compared to other methods, the improved Hopf oscillator can realize various gait patterns and achieve robust and controllable transitions.

For quadruped locomotion, the CPG can be represented by four coupled Hopf oscillators in the following way\cite{7090642}
\begin{equation}
	\label{eq_method_hopf_1}
	\dot{\boldsymbol{\rho}}_i
	=
	f\left(\boldsymbol{\rho}_i\right)
	+
	\sum_{j}{R_{ij}\boldsymbol{\rho}_i}
\end{equation}
where $f\left(\boldsymbol{\rho}_i\right)$ represents the system in (\ref{eq_method_hopf_0}), and $\delta$ represents the coupling strength and the matrix ${R}_{ij}$ is the connection matrix that defines the coupling pattern of four oscillators. The connection matrix defines the foot sequence, which is the major gait character. The connection matrix can be formulated in the following form
\begin{equation}
	{R}_{ij}=
	\begin{bmatrix}
		\cos \theta_{{\rm d}ij} & -\sin \theta_{{\rm d}ij} \\
		\sin \theta_{{\rm d}ij} & \cos \theta_{{\rm d}ij}
	\end{bmatrix}
\end{equation}
where $\theta_{{\rm d}ij}=\varphi_{{\rm d}i}-\varphi_{{\rm d}j}$ is the desired difference between oscillator $i$ and oscillator $j$; $\varphi_{{\rm d}i}$ is the desired relative phase, in comparison to the actual phase $\varphi_i$. The coupling term in (\ref{eq_method_hopf_1}) is critical in keeping the phase difference between oscillator $i$ and $j$.

The Hopf oscillator CPG is powerful in gait generation. Table. \ref{method_table_1} lists parameters of the period, duty factor and desired phase offset, that realize gaits of trot, pacing, bounding and walk, which are four common gaits in quadruped animals at different speed\cite{Xi2016,Zhang2018}. Other parameters are $\alpha=50,\mu=1,b=50,\delta=1$. A three-legged walk can also be generated by simply placing a fixed attractor, \textit{i.e.}, $\varphi=-\frac{\pi}{2},r=1$, in the Hopf oscillator that generates phase of the holding leg. The oscillator will approach the attractor asymptotically and remain pinned. Consequently, the leg will stay at the highest point of the swing phase. By placing the attractor to a different oscillator, one can change the holding leg during locomotion.
\begin{table}[!t]
	\renewcommand{\arraystretch}{1.3}
	\caption{CPG Parameters for Various Gaits}
	\label{method_table_1}
	\centering
	\begin{tabular}{|c|c|c|c|}
		\hline
		Gait & $T$ & $\beta$ & $\boldsymbol{\hat{\varphi}}$\\
		\hline
		Trot & $0.5s$ & $0.5$ & $\left[0, \pi, \pi, 0\right]$\\
		\hline
		Pacing & $0.5s$ & $0.5$ & $\left[0, \pi, 0, \pi\right]$\\
		\hline
		Bounding & $0.3s$ & $0.4$ & $\left[0, 0, \pi, \pi\right]$\\
		\hline
		Four-legged Walk & $0.6s$ & $0.75$ & $\left[0, 0.5\pi, \pi, 1.5\pi\right]$\\
		\hline
		Three-legged Walk & $0.45s$ & $2/3$ & $\left[0, 2\pi/3, 4\pi/3\right]$\\
		\hline
	\end{tabular}
\end{table}

As the foot sequence is defined by the connection matrix, the gait transition can be realized by simply replacing one matrix with another. Generally, the transition speed is controlled by the parameter $\delta$ in (\ref{eq_method_hopf_1}). As the coupling becomes stronger, the transition becomes faster. To achieve stable transition, the transition phase offset ${\theta}^{trans}_{\rm d}$ is the summation of the desired phase offset ${\theta}^{target}_{\rm d}$ for the target gait and an overshot term\cite{7090642} in (\ref{eq_method_overshoot}).
\begin{eqnarray}
	\label{eq_method_overshoot}
	\begin{aligned}
		{\theta}^{trans}_{\rm d}&={\theta}^{{\rm d}target}_{\rm d} +
		k\eta \left({\theta}^{target}_{\rm d} - {\theta}^{init}_{\rm d}\right) \\
		\eta&=\frac{\theta_{{\rm d}mn}^{target}-\theta_{mn}^{current}}{{\theta}_{{\rm d}mn}^{target}-{\theta}_{{\rm d}mn}^{init}}
	\end{aligned}
\end{eqnarray}
In (\ref{eq_method_overshoot}), ${\theta}^{init}_{\rm d}$ is the designed phase offset of the initial gait; ${\theta}^{current}$ is the actual phase offset used during the gait transition; $k$ is a positive constant representing the strength of the overshoot; $\eta$ represents the remaining progress in transition; $\theta_{mn}$ is the selected phase difference to indicate the progress. The overshoot let $\theta^{trans}_{\rm d}$ start from ${\theta}^{{\rm d}target}_{\rm d}+k\left({\theta}^{target}_{\rm d} - {\theta}^{init}_{\rm d}\right)$ and decrease to $\theta^{target}_{\rm d}$.It is designed to avoid some dead zone during the transition process and accelerate the transition. The differential equation of the system ensures smooth and robust transition process regardless of the starting time of transition.

\begin{figure}[!t]
	\centering
	\includegraphics[width=3.4in]{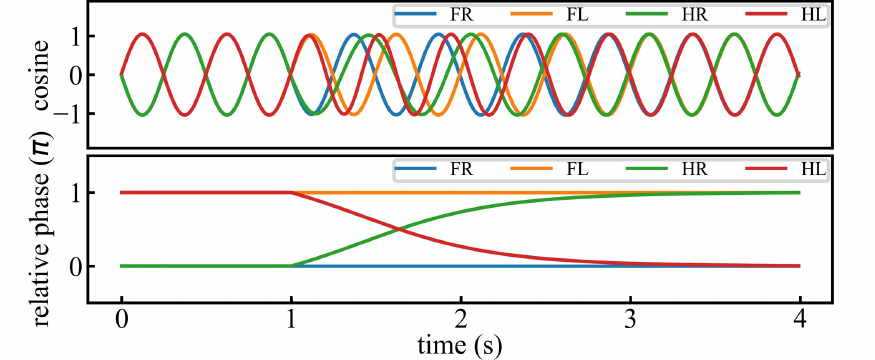}
	\caption{Illustration for phases during the transition from pace to trot. The transition process starts from 1.0s and converge at 2.48s. The upper figure illustrates the cosine values of absolute phases while the lower figure illustrates the relative phases. FR, FL, HR and HL represent the right front, the left front, the right hind and the left hind legs, respectively.}
	\label{method_5}
\end{figure}

Fig. \ref{method_5} shows the transition from pacing to trot. We use the cosine value and relative differences of phases to show the transition state. The overshoot coefficient $k$ is set to $0.5$ in this work. The transition starts at $t=1.0s$, and the overshoot takes the maximum effect with $\eta=1.0$. The relative phase offset changes rapidly at the beginning time. The oscillator for the hind left leg speeds up while the oscillator for the front left leg slows down. At $t=2.0s$, the remaining process is $\eta=0.26$, and the transition becomes slower and smoothly converges to the new gait. At $t=2.48s$, the remaining process is $\eta=0.1$, and we regard this point to be the converge point. 

The reference motion can also be generated by using manually designed periodic functions in the form of $f:\left[0, +\infty\right)\rightarrow \left[-\pi,\pi\right)$, which provides more flexibility. For example, dancing steps can be created according to the rhythms of the music. Equation (\ref{eq_method_manual}) shows an example in which one diagonal pair of legs moves in a period two-times longer than the other pair. 
\begin{eqnarray}
	\label{eq_method_manual}
	\begin{aligned}
		\varphi_{1,4}\left(t\right)&=
		\begin{cases}
			4\pi t, &t\in[0, 0.5),\\
			\varphi_{1,4}\left(t-0.5\right), &t\in[0.5, +\infty)
		\end{cases} \\
		\varphi_{2,3}\left(t\right)&=
		\begin{cases}
			2\pi t, &t\in[0, 1),\\
			\varphi_{2,4}\left(t-1.0\right), &t\in[1, +\infty)
		\end{cases} \\
	\end{aligned}
\end{eqnarray}

\subsection{Control Architecture}
The control architecture is schematically shown in Fig. \ref{method_1}. The controller input $\boldsymbol{X}=\left\{\boldsymbol{v}^{des}, \boldsymbol{t}, \boldsymbol{q}_j, \dot{\boldsymbol{q}}_j, \dot{\boldsymbol{q}}_{b_R}, \boldsymbol{r}_g\right\}\in\mathbb{R}^{41}$ can be divided into three categories. The first part is the velocity command from user. $\boldsymbol{v}^{des}\in\mathbb{R}^3$ is the desired velocities in forward, lateral and yaw direction. The second part is the sine and cosine values of four phases $\boldsymbol{t}\in\mathbb{R}^8$. Benefit from our parameterization to the reference motion, we can use phases as the input when imitating multiple reference motions, instead of the several steps reference motion itself. The last part is the state of the robot. $\boldsymbol{q}_j\in\mathbb{R}^{12},\dot{\boldsymbol{q}}_j\in\mathbb{R}^{12}$ are the position and velocity of all 12 joints, respectively; $\dot{\boldsymbol{q}}_{b_R}\in\mathbb{R}^{3}$ represent the angular velocities obtained directly from the inertial measurement unit, the unit vector $\boldsymbol{r}_g$ indicates the gravity direction in the base frame, representing the angular position of the base. No further state estimation is used in this controller. The output of the controller is the target position $\boldsymbol{q}^{target}_j\in\mathbb{R}^{12}$ of each joint. The low-level torque command to each motor is determined by a PD controller. We take the long short-term memory (LSTM) neural network as the RL controller, which which can adaptively encode the historical information and has been demonstrated to have advantage over naive multilayer perceptron  (MLP) when dealing with dynamic processes\cite{siekmann2020learning}.

\subsection{Reward Design}
The policy is trained to imitate reference motion while maintaining balance and tracking command velocities. The reward function is a weighted summation of separate terms in the following form:
\begin{eqnarray}
	\begin{aligned}
		r=&\boldsymbol{w}\boldsymbol{r}^o\\
		\boldsymbol{r}^o=&\left[r_\tau,r_v,r_j,r_b,r_h,r_c\right]^T \\
		\boldsymbol{w}=&\left[0.2,0.15,0.5,0.05,0.05,0.05\right] \\
		r_{j}=&0.25\exp{\left(-2\lVert\Delta \boldsymbol{q}_j\rVert^2\right)}
		+ 0.75\exp{\left(-2\lVert\Delta \dot{\boldsymbol{q}}_j\rVert^2\right)} \\
		r_{\tau}=&0.5\exp{\left(-\left\lVert0.05\boldsymbol{\tau}\right\rVert^2\right)}
		+0.5\exp{\left(-\left\lVert0.5\dot{\boldsymbol{\tau}}\right\rVert^2\right)}\\
		r_{v}=&\exp{\left(-8\lVert\Delta \dot{\boldsymbol{q}}_b\rVert^2\right)} \\
		r_{b}=&\exp{\left(-80\lVert \boldsymbol{r}_{g}-\boldsymbol{z}\rVert^2\right)} \\
		r_{h}=&\exp{\left(-80\lVert q_{b_z}-q_{b_z}^{des}\rVert^2\right)} \\
		r_{c}=&\exp{\left(-\sum_{i}{\mathbb{F}_i\sin^2{\varphi_i}}\right)} \\
		\mathbb{F}_i=&c_i\lVert 0.08\boldsymbol{f}_i\rVert^2+(1-c_i)\lVert2\boldsymbol{v}_i\rVert^2
	\end{aligned}
\end{eqnarray}
in which $r_\tau$ penalizes the policy for large or rapid joint torque; $r_{v}$ rewards the policy for following the desired velocity; $r_{b}$ and $r_h$ reward the policy for keeping the base balance; $\boldsymbol{z}$ is the unit vector along the z-axis in the base frame; $r_c$ penalizes the policy for tracking the foot contact state of reference motion. $c_i$ represents the desired contact state of the i-th foot. $\boldsymbol{f}_{fi}$ is the contact force on the $i$-th foot and $\boldsymbol{v}_{fi}$ the velocity. $c_i=0$ for the stance state and $c_i=1$ for swing state. When the $i$-th foot is supposed to be in the swing state, $c_i=1$ and only the force on the $i$-th foot will decrease the reward. On the other hand, when the $i$-th foot is supposed to be in the stance state, $c_i=0$ and only the velocity of the $i$-th foot in the world frame will decrease the reward. The term $\sin^2{\varphi_i}$ is designed to reduce the effect of this penalty at $\varphi_i=0$ and $\pi$ when the desired contact state changes. This reward helps to keep the feet in the correct phases and avoid slipping. The term $r_j$ represents the major part of imitation and rewards the policy for tracking both joint positions and velocities of the reference motion. The imitation reward by integrating the reference motion is general to all gaits and serves as a soft constraint for the exploration of the agent to avoid learning unnatural behaviors, including the fusion of multiple gaits. This reward has a large coefficient of $0.5$ to encourage the policy to track the reference motion while following the target velocities and retaining balance.

\subsection{Training Techniques}
Domain randomization can help to overcome the gap between simulation and real world and to prevent overfitting\cite{Hwangboeaau5872,siekmann2020learning}. To minimize the sim-to-real gap, we randomize the dynamics and kinematics parameters of the robot as well as the ground. Small but random external forces are applied to the robot. To model the sensory noise, we also apply randomization to the sensory data in the observation. For faster convergence, we adopt the reference state initialization (RSI) technique, proposed by Peng \textit{et al}.\cite{2018-TOG-deepMimic} To expedites ergodicity, the initial state is randomly sampled from the reference motion. Otherwise, the policy NN may depend more heavily on a portion of the states if the initial state is fixed in the training process. Details on the random variables used in this part can be found in Table \ref{method_table_2}.
\begin{table}[!t]
	\renewcommand{\arraystretch}{1.3}
	\caption{Randomization During Training}
	\label{method_table_2}
	\centering
	\begin{tabular}{|c|c|c|}
		\hline
		Parameter & Unit & Distribution\\
		\hline
		External Force & N & $\mathcal{U}\left(0, 10\right)$\\
		\hline
		External Torque & Nm & $\mathcal{U}\left(0, 2\right)$\\
		\hline
		Ground Friction & - & $\mathcal{U}\left(0.4, 1.2\right)$\\
		\hline
		Ground Height & mm & $\mathcal{U}\left(-2.5, 2.5\right)$\\
		\hline
		Mass & kg & $\mathcal{N}\left(1.0, 0.05\right)\times$default values\\
		\hline
		Body Size & m & $\mathcal{N}\left(1.0, 0.05\right)\times$default values\\
		\hline
		Joint Position Noise & rad & $\mathcal{N}\left(0.0, 0.002\right)$\\
		\hline
		Joint Velocity Noise & rad/s & $\mathcal{N}\left(0.0, 0.3\right)$\\
		\hline
		Body Posture Noise & rad & $\mathcal{N}\left(0.0, 0.1\right)$\\
		\hline
		Angular Velocity Noise & rad & $\mathcal{N}\left(0.0, 0.3\right)$\\
		\hline
	\end{tabular}
\end{table}

\section{RESULTS}
To evaluate our learning framework, we deploy the control policy on the Black Panther robot, a medium-dog-size quadruped robot with 12 degrees-of-freedom. The control policy shows similar behaviors in both simulation and hardware. In the following experiments, we first demonstrate the effectiveness for learning multiple gaits. We then show the performance of present method in gait transitions and more complex gaits. Overall, the controller is robust to different ground conditions and unknown perturbation in the real world as shown in Supplementary video 1.

\subsection{Experiment Setup}
RaiSim\cite{raisim} is used as the physics engine for simulation. The control policy is an LSTM neural network with 2 hidden layers of 128 units in each, trained with PPO algorithm\cite{schulman2017proximal} implemented in the stable-baselines\cite{stable-baselines} package. To balance the convergence speed and the exploration field, we adopt a truncated trainable action noise to for exploration with standard deviation initialized at 1.0 and ended with 0.1. During the sampling process, the controller works at a frequency of 100Hz in simulation while the physical engine updates at 400Hz. Each batch contains 200 trajectories up to 5 seconds in simulation, corresponding to 500 control steps. All the gait patterns, including walk, trot, pace, bounding, three-legged and dancing patterns, are trained together within one single policy. The RL environment is released as an open-source package.\footnote{https://github.com/ZJU-XMech/PhaseGuidedControl} The policy is trained on a workstation with intel i9-10900X CPU and RTX 2080Ti GPU and converges in about 3 hours.

The Black Panther robot is designed based on the open-source project, mini cheetah\cite{katz2019}. The base of the robot is about 27 cm wide and 42 cm long. The max length of each leg is about 39 cm, and the total weight is about 9 kg. Each leg has the abductor, hip, and knee joints. We use the UP Board as the onboard computer, on which the policy can run at 100Hz as designed.

\subsection{Multiple Gait Locomotion}\label{sec_result_gait}
\begin{figure}[!t]
	\centering
	\includegraphics[width=3.4in]{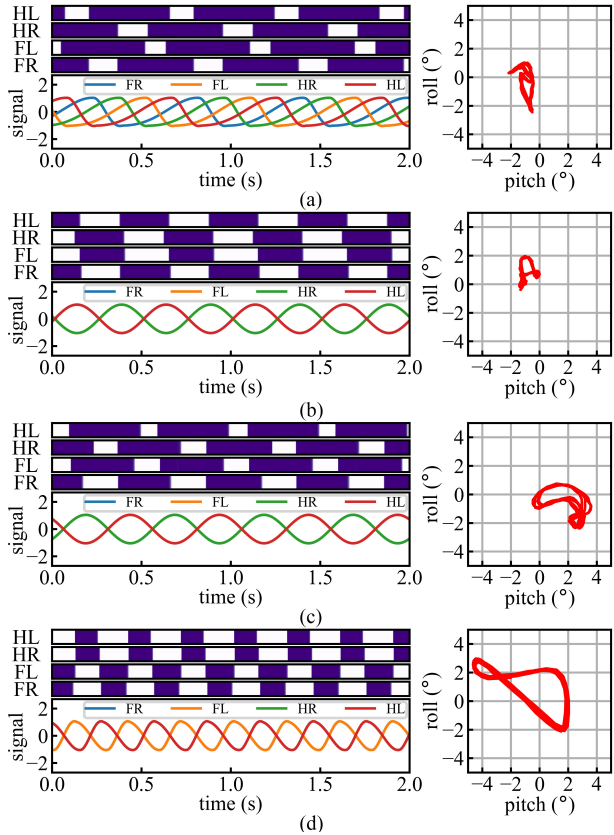}
	\caption{Simulation of four common gaits including (a) walk, (b) trot, (c) pace and (d) bounding. The actual foot contact sequence is represented in the horizontal bars. The CPG output is plotted with the cosine value of the phase, and thus the rising part stands for the stance state and the falling part the swing phase.}
	\label{result_1}
\end{figure}

We evaluate the control performance for four most common gaits, namely walk, trot, pacing, and bounding. Parameters for those four gaits are specified in Table \ref{method_table_1}. Fig. \ref{result_1} shows the simulated results, and Table \ref{result_table} lists the detailed metrics for the tracking of velocities and contact sequence measured in $2s$. The corresponding deployment on the hardware is provided in the supplementary video 2. The horizontal bars represent the actual foot contact sequence. Dark bars stand for the stance state, and the white bars the swing state. For all four gaits, the foot sequence can match the given CPG output and the step height error is within 2 cm. Since the policy is encouraged to reduce joint torque, the actual swing height is generally lower than designed. We further measure the similarity between the desired and actual contact sequence by $s=1-d/T$, in which $d$ stands for the Hamming distance of two time series and $T$ stands for the length of the sampled time series. Since the policy is not trained to control the roll and pitch angle of the base according to user command, it is trained to keep the base balance, and thus we use the roll and pitch angles as a measurement of balance. For the walk gait, each foot is supposed to step forward in one quarter period. The roll and pitch angles of the robot are within $2.5^{\circ}$. The robot stays stable without any trend of losing balance. For the trot gait, the feet on the diagonal has the same phase value. The roll and pitch angles of the robot are within $2.0^{\circ}$. For the pacing gait, feet on the same side in the lateral direction are supposed to move in the same phase. The foot sequence is able to match the CPG output but the duration of the stance state is $50\%$ longer than the expectation, because maintaining balance is hard when both two legs on the same side are in the swing state. The agent learns to increase the duration of the stance state. The roll and pitch angles of the robot are within $4.0^{\circ}$. For the bounding gait, feet on the same side in the front and back direction are supposed to move in the same phase. The robot can maintain balance. However, the pitch angle is larger than other three gaits, because the frequency of bounding is higher and lifting both two legs on the same side in the front and back direction would naturally lead to larger pitch angle.
\begin{table}[!t]
	\renewcommand{\arraystretch}{1.3}
	\caption{Performance Metrics}
	\label{result_table}
	\centering
	\begin{tabular}{|c|c|c|}
		\hline
		Gait & Similarity & Velocity Error ($m/s$)\\
		\hline
		Walk & 0.94 & 0.04 \\
		\hline
		Trot & 0.95 & 0.12 \\
		\hline
		Pacing & 0.78 & 0.15 \\
		\hline
		Bounding & 0.87 & 0.16 \\
		\hline
	\end{tabular}
\end{table}

\subsection{Gait Transitions}
\begin{figure*}[!t]
	\centering
	\includegraphics[width=7.0in]{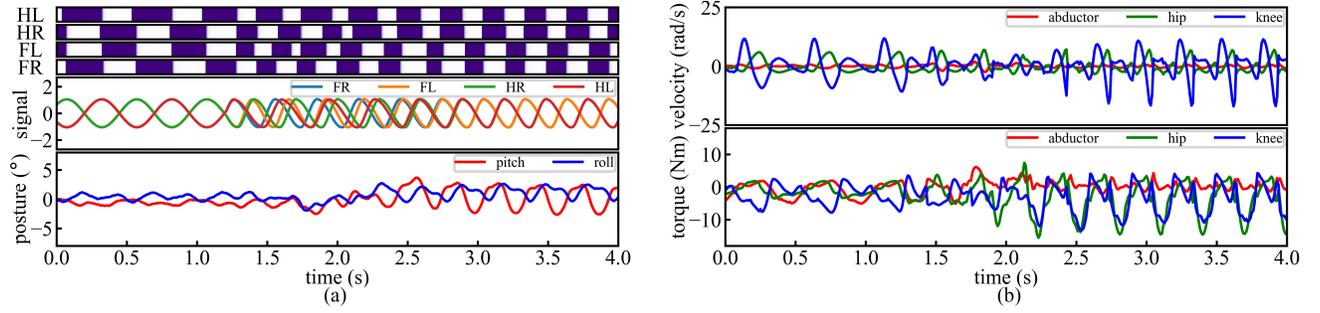}
	\caption{A typical gait transition from trot to pacing. (a) The actual foot fall, CPG signals and the roll and pitch angles during the transition process. (b) The maximum speed and torque for three types of joint indicates that the transition process is smooth at joint level.}
	\label{result_2}
\end{figure*}

We test the ability for gait transition in this section. Parameters used in this section are specified in Section \ref{sec_gait}. Supplementary video 2 shows the hardware test of all the transitions among four gaits tested above. Fig. \ref{result_2} shows the transition from trot to bounding at a forward velocity of 0.5 m/s. The CPG signals show that the transition starts at about $t=1.2s$ and finishes at about $t=2.5s$. The actual foot sequence can match the CPG output during the transition process, with a similarity of 0.83. The velocity tracking error during transition is $0.21m/s$. The body postures before and after the transition are similar to the results in Section \ref{sec_result_gait}. During the transition process, the robot maintains its balance, and the amplitudes of the roll and pitch angles change smoothly. Since the guiding signal from CPG is smooth during transition, the reference motion learned by the controller is smooth. Eventually the smooth working condition of motors is achieved, and no peak or sudden change occurs during the transition process.

\subsection{Three-legged Gaits}
\begin{figure*}[!t]
	\centering
	\includegraphics[width=7.0in]{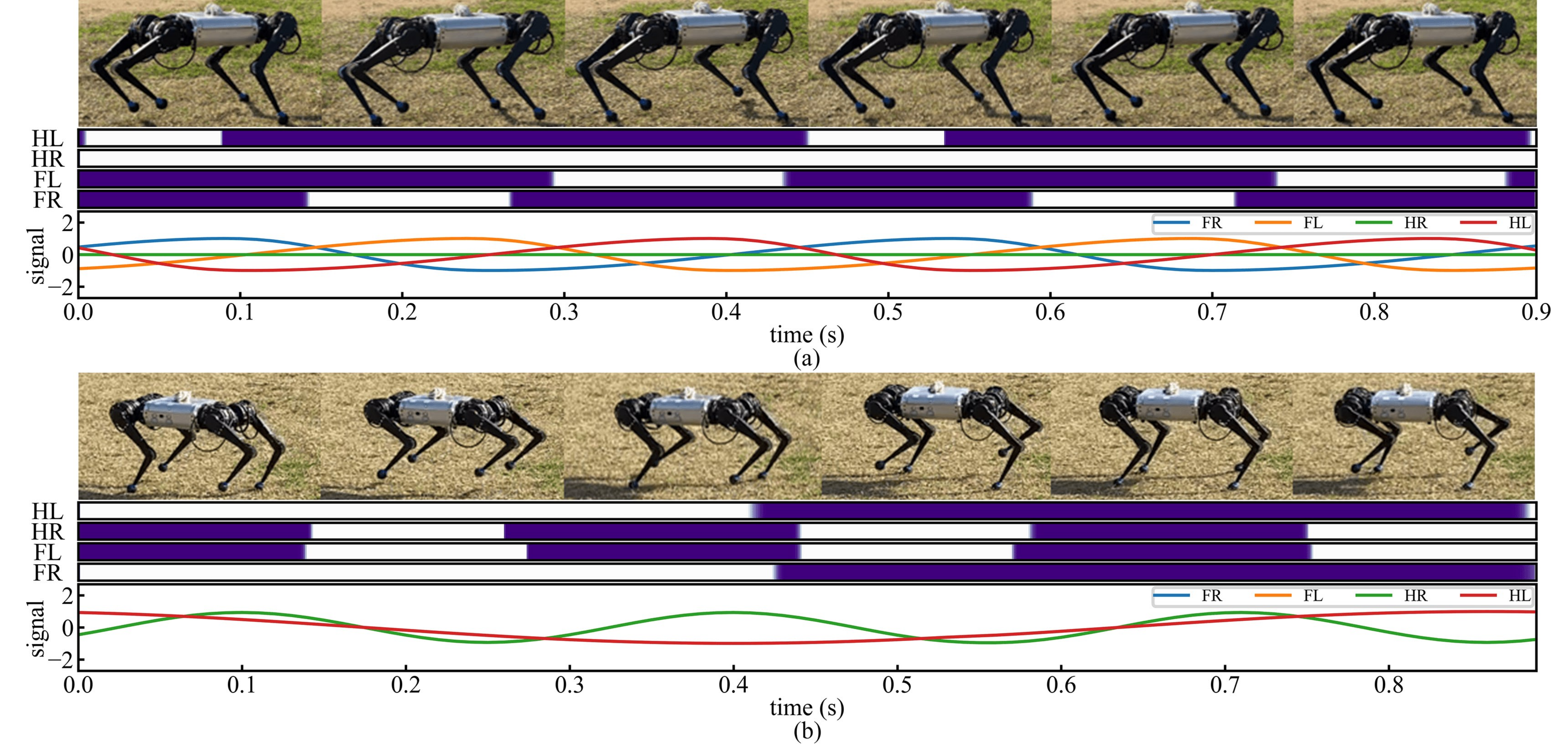}
	\caption{Examples for complex gait deployed on the hardware. (a) An example of three-legged locomotion. (b) An example of a manually designed gait in which four legs hold various periods. Each foot is found to be in the proper swing or stance phase as expected.}
	\label{result_3}
\end{figure*}

The three-legged locomotion and transition are also tested and deployed on the hardware, as shown in Fig. \ref{result_3} (a). The parameters are specified in Table \ref{method_table_2}. The deployment details are provided in the Supplementary video 3. The robot can walk with three legs following velocity command, and switch the holding leg during walking. The foot sequence can match the CPG signals even during the switch process. After the switch, the robot can reach steady state in less than $1$ second. 

\subsection{Complex Gaits}
\begin{figure}[!t]
	\centering
	\includegraphics[width=3.4in]{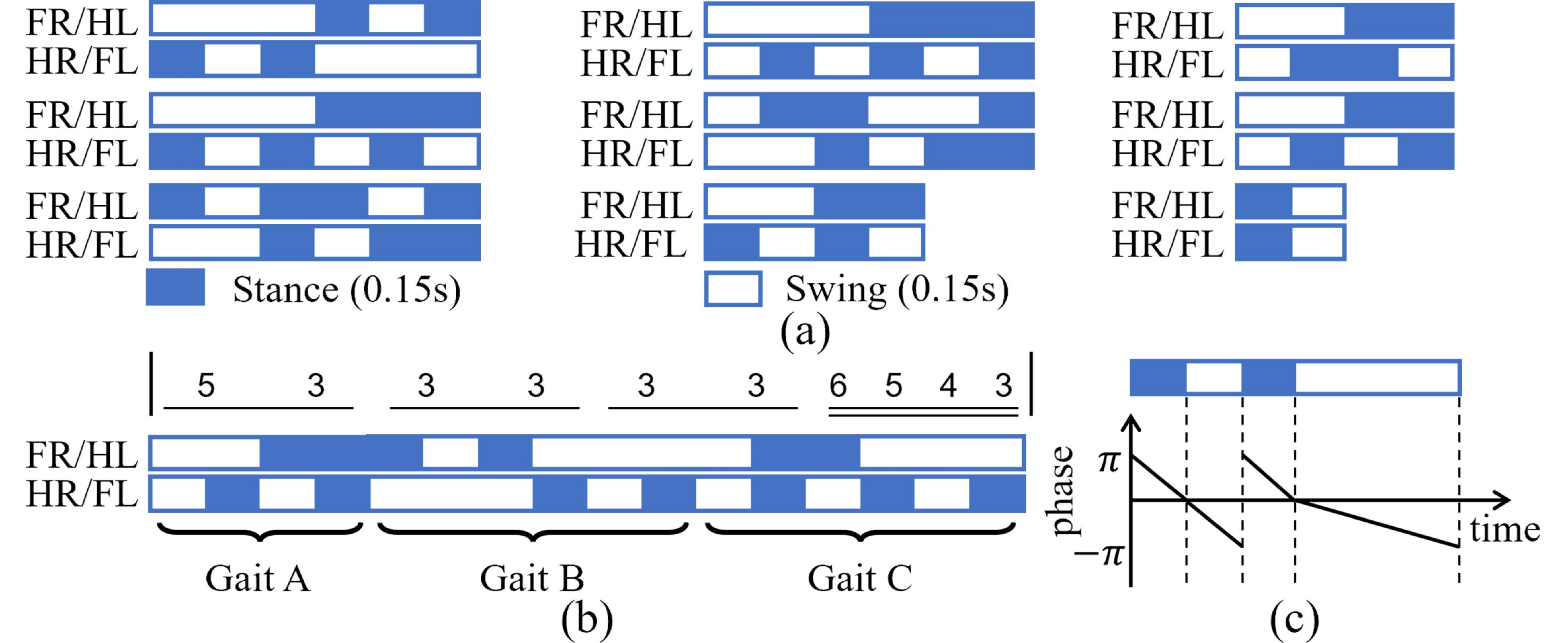}
	\caption{(a) Illustration for dancing steps. The foot sequence for each gait in one period is shown as horizontal bars. The length of the bars represents the length of duration. (b) Mapping from the notation of the rhythm to the foot sequence. (c) Relationship between the foot sequence and the phase.}
	\label{result_4}
\end{figure}

Dancing gaits can be generated by manually designing periodic functions according to the rhythms of the music. As a demonstration, we use the music from \textit{Dances of the Swans}, the fourth composition of the second act in the ballet \textit{Swan Lake}, created by Tchaikovsky. The tempo used in this experiment is 100 bpm. We define the unit time for a swing or stance duration as $t_0=0.15s$, equal to a quarter of one beat in the music, the same speed as the ballet. Various kinds of rhythmic gaits can be designed in this way. To avoid tediously listing a large number of similar equations, we show the foot sequence of those gaits in horizontal bars in Fig. \ref{result_4} (a). Fig. \ref{result_4} (b) shows an example of the mapping from the rhythm to the foot sequence. Fig. \ref{result_4} (c) shows the relationship between the foot sequence and the phase. The corresponding function for Fig. \ref{result_4} (c) is shown in (\ref{eq_result_manual}).
\begin{eqnarray}
	\label{eq_result_manual}
	\begin{aligned}
		\varphi\left(t\right)&=
		\begin{cases}
			-\frac{\pi}{0.15} t + \pi, &t\in[0, 0.3),\\
			-\frac{\pi}{0.15} (t-0.3) + \pi, &t\in[0.3, 0.45), \\
			-\frac{\pi}{0.45} (t-0.45), &t\in[0.45, 0.9), \\
			\varphi\left(t-0.9\right), &t\in[0.9, +\infty)
		\end{cases} \\
	\end{aligned}
\end{eqnarray}

Those 9 gaits contain various dynamic situations, like jumping and landing with different feet, moving different legs at different and changing frequency, \textit{etc}. The controller is able to distinguish all those gaits according to the phases and adapt to all the working conditions. To transit from CPG-based gaits to manually designed gaits, we first adjust the phase offset of the CPG and then replace it with manually designed functions when the exact phases evolve to the state that exist both in two generators, and vice versa. Eventually the robot can walk in all those 9 gaits and transit freely from one to another while following the velocity commands and phases. Outdoor tests for all 9 gaits are shown in the supplementary video 4. The one-minute dancing is provided in the supplementary video 5. Simulation results for transition between CPG-based gaits and manually designed gaits is provided in the supplementary video 6.

\section{CONCLUSION}
In this work, we present a reinforcement learning framework to train a phase-guided controller for multiple gaits and free transitions. We use phases as the interface between gait generator and the RL system. For the control policy, the learning task is simplified, and the agent only needs to learn the relationship between phases and foot movements while retaining balance. Eventually, the policy trained in simulation can be transferred to hardware and reproduce all the learned gaits and free transitions robustly.

This framework transforms multiple gait learning to learning the relationship between phases and foot movements, which avoids the challenge of multi-task learning problem. In this way, the LSTM neural network is sufficient to learn those locomotion skills. The deployment on the hardware has demonstrated that the robot can locomote in various gaits and make free transitions according to the guided phases. Under this framework, the robot locomotion is closely related to the gait generation, which is a much simpler problem. By designing more advanced gait generators, the gap between legged robots and animals can be narrowed and eventually overcome.

\section*{ACKNOWLEDGMENT}
This work was supported by the State Key Laboratory of Fluid Power and Mechatronic Systems (Zhejiang University).

\bibliographystyle{IEEEtran}
\bibliography{bibtex/bib/reference.bib}

\end{document}